\documentclass[
]{ceurart}

\usepackage{listings}
\usepackage[T1]{fontenc}
\usepackage{graphicx}
\usepackage{booktabs}
\usepackage{booktabs}
\usepackage{pifont}
\usepackage{float}
\usepackage{tabularx}   % automatic width for text columns
\usepackage{xcolor}
\usepackage{comment}
\usepackage{array}
\usepackage{csquotes}
\usepackage{subcaption}
\usepackage[most]{tcolorbox}
\begin{document}

\copyrightyear{2025}
\copyrightclause{Copyright for this paper by its authors.
  Use permitted under Creative Commons License Attribution 4.0
  International (CC BY 4.0).}

\conference{CLEF 2025 Working Notes, September 9– 12 September 2025, Madrid, Spain}

\title{MSA at ImageCLEF 2025 Multimodal Reasoning: Multilingual Multimodal Reasoning With Ensemble Vision-Language Models}

\tnotemark[1]

\author[1]{Seif Ahmed}[%
email=seifeldein.ahmed@msa.edu.eg,
]
\cormark[1]
\fnmark[1]
\address[1]{October University for Modern Sciences and Arts (MSA),
  Giza, Egypt}

\author[1]{Mohamed T. Younes}[%
email=mohamed.tarek61@msa.edu.eg,
]
\fnmark[1]

\author[1]{Abdelrahman Moustafa}[%
email=abdelrahman.moustafa5@msa.edu.eg,
]
\fnmark[1]

\author[1]{Abdulrahman Allam}[%
email=abdulrahman.atif5@msa.edu.eg,
]
\fnmark[1]

\author[1]{Hamza Moustafa}[%
email=hamza.moustafa@msa.edu.eg,
]
\fnmark[1]

\cortext[1]{Corresponding author.}
\fntext[1]{These authors contributed equally.}

\begin{abstract}
We present a robust ensemble-based system for multilingual multimodal reasoning, designed for the ImageCLEF 2025 EXAMS-V challenge. Our approach integrates Gemini 2.5 Flash for visual description, Gemini 1.5 Pro for caption refinement and consistency checks, and Gemini 2.5 Pro as a reasoner which handles final answer selection, all coordinated through carefully engineered few-shot and zero-shot prompts. We conducted an extensive ablation study, training several large language models (Gemini 2.5 Flash, Phi-4, Gemma-3, Mistral) on an English dataset and its multilingual augmented version. Additionally, we evaluated Gemini 2.5 Flash in a zero-shot setting for comparison and found it to substantially outperform the trained models. Prompt design also proved critical: enforcing concise, language-normalized formats and prohibiting explanatory text boosted model accuracy on the English validation set from 55.9\% to 61.7\%. On the official leaderboard, our system (Team MSA) achieved first place overall in the multilingual track with 81.4\% accuracy, and led 11 out of 13 individual language tracks, with top results such as 95.07\% for Croatian and 92.12\% for Italian. These findings highlight that lightweight OCR–VLM ensembles, when paired with precise prompt strategies and cross-lingual augmentation, can outperform heavier end-to-end models in high-stakes, multilingual educational settings.
\end{abstract}


\maketitle

\section{Introduction}

Vision-Language Models (VLMs) have rapidly advanced in recent years, demonstrating remarkable capabilities in diverse multimodal tasks such as image captioning, visual question answering (VQA), and visual dialogue~\cite{zhang2024vision,wang2024exploring}. Despite these successes, contemporary VLMs often encounter significant challenges in tasks demanding deep logical reasoning or inferencing~\cite{bi2025reasoning,lu2022learn}. Limitations in the current generation of models are frequently revealed by complex queries involving intricate dependencies or hypothetical scenarios. Thus, it remains crucial to rigorously assess how well modern language and vision models can reason across complex multimodal inputs, especially in multilingual contexts~\cite{li2024m4u,huang2023m3exam,gao2025pm4benchparallelmultilingualmultimodal}. For a detailed description of the shared task and competition, we refer the reader to the official overview papers~\cite{ImageCLEFmultimodalReasoningOverview2025,OverviewImageCLEF2025}.

To address these challenges, three distinct tasks have emerged to evaluate VLM performance across various reasoning scenarios. Task 1, Visual Question Answering (VQA), requires the generation of accurate textual answers from image-question pairs, demanding precise interpretation and description of image content~\cite{bi2025reasoning,zhang2024vision}. Task 2, Visual Question Generation (VQG), involves generating contextually relevant questions from given images and associated answers, testing models' ability to deeply understand visual contexts and formulate meaningful textual queries~\cite{lu2022learn}. Task 3, Visual Location Question Answering (VLQA), further extends these challenges by requiring spatial localization through segmentation masks, necessitating accurate identification and delineation of regions of interest based on textual prompts. Our task was focused only on the Visual Question Answering (VQA) task. 

Motivated by the complexities and novel demands of recent multimodal reasoning benchmarks~\cite{gao2025pm4benchparallelmultilingualmultimodal,li2024m4u,huang2023m3exam}, our approach leverages a strategic ensemble of advanced transformer-based models, specifically integrating Gemini 2.5 Flash for enhanced visual understanding and Gemini 1.5 Pro coupled with Gemini 2.5 Pro for sophisticated reasoning and answer aggregation. This hybrid approach exploits the complementary strengths of each model, achieving robust performance across multilingual datasets.

Our contributions in this paper are threefold: First, we provide a detailed examination of our system's architecture and the rationale behind model selection and combination. Second, we thoroughly analyze the performance of our system on multilingual multimodal reasoning tasks, emphasizing insights gained from multilingual diversity and complexity. Finally, we reflect on lessons learned from the evaluation and suggest pathways for future enhancements to strengthen multimodal reasoning capabilities.

\section{Related Work}

Recent advancements in multimodal and multilingual reasoning have underscored the complexity and richness of these domains. Benchmarks such as M4U~\cite{li2024m4u,wang2025m4uevaluatingmultilingualunderstanding}, M3Exam~\cite{huang2023m3exam,NEURIPS2023_117c5c86}, and PM4Bench~\cite{gao2025pm4benchparallelmultilingualmultimodal} have emerged as pivotal platforms for evaluating large multimodal models across diverse languages and complex reasoning tasks. These benchmarks facilitate rigorous assessment of model capabilities in multilingual understanding, multimodal reasoning, and multi-level inference, encompassing various modalities such as text, images, and video.

The reasoning ability of language models, especially via chain-of-thought prompting, has also been extensively explored and shown to be particularly effective in multilingual contexts~\cite{shi2022languagemodelsmultilingualchainofthought,zhou2022language}. This research emphasizes the necessity of developing robust models capable of handling multilingual data and highlights the benefits of incorporating explicit reasoning steps within model architectures. Recent large models like GPT-4 and Gemini have demonstrated significant progress in multilingual reasoning, maintaining logical coherence across diverse linguistic settings.

Multimodal reasoning tasks such as Visual Question Answering (VQA), Visual Question Generation (VQG), and Visual Location Question Answering (VLQA) have notably benefited from transformer-based architectures and vision-language model innovations~\cite{zhang2024vision,lu2022learn}. Techniques including Vision Transformers (ViT), SegFormer, and VisualBERT have shown promising results in interpreting visual information and generating relevant textual content. These transformer-based models leverage self-attention mechanisms to integrate visual and textual features, facilitating a nuanced understanding of multimodal inputs~\cite{li2024m4u,huang2023m3exam}.

Recent research also highlights the role of evaluation methodologies and metrics in accurately capturing model performance~\cite{bi2025reasoning,wang2024exploring}. Evaluations commonly include metrics such as accuracy, precision, recall, Intersection-over-Union (IoU), and Dice coefficients especially for tasks involving segmentation masks. The increasing complexity of multimodal tasks necessitates advanced evaluation strategies, as discussed in recent benchmarks, which systematically categorize challenges in visual question answering and generation, and underscore the importance of precise metrics to evaluate nuanced performances~\cite{ImageCLEFmultimodalReasoningOverview2025,OverviewImageCLEF2025}.

Collectively, these studies underscore the ongoing need for sophisticated models capable of intricate multimodal reasoning, highlighting both the progress made and the challenges remaining in the field. Continued research and development are essential to addressing existing limitations and unlocking further advancements in multimodal and multilingual reasoning capabilities.

\section{Dataset and Task Description}

It is shown through Table~\ref{tab:dataset_stats}, that the multilingual dataset under study consists of over 20{,}000 questions distributed across 13 languages including English, Chinese, German, Spanish, Arabic, Italian, Bulgarian, Croatian, Serbian, Urdu, Polish, and Kazakh. Each question is associated with metadata such as sample\_id, subject (e.g., biology, chemistry, physics), type (text or image\_text), grade (ranging from 4 to 12), answer\_key (A, B, C, D, or E), and language [as shown in Table~\ref{tab:dataset_stats}]. The questions span a variety of educational domains and cognitive skills, presenting a comprehensive challenge for multimodal reasoning systems.

\begin{figure*}[h]
  \centering
  \begin{subfigure}[t]{0.31\textwidth}
      \includegraphics[width=\linewidth]{Images/Arabicinenglishdataset_resized.png}% ← 1st image
      \caption{Example of answer options entirely in \textbf{Arabic} although the metadata tag says “English”.}
  \end{subfigure}\hfill
  \begin{subfigure}[t]{0.31\textwidth}
      \includegraphics[width=\linewidth]{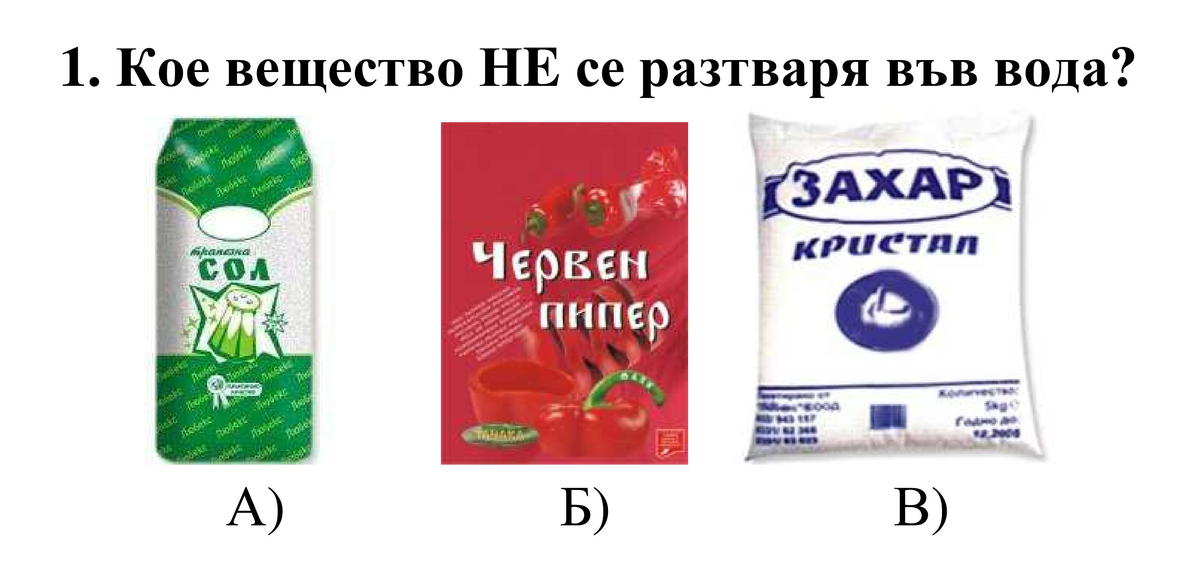}% ← 2nd image
      \caption{Example of answer options labeled in \textbf{Bulgarian letters} which the OCR fails to map to \{A,B,C,D,E\}.}
  \end{subfigure}\hfill
  \begin{subfigure}[t]{0.31\textwidth}
      \includegraphics[width=\linewidth]{Images/nolabels_resize.png}% ← 3rd image
      \caption{Example of answer options completely \textbf{unlabeled}.}
  \end{subfigure}
  \caption{Illustrative OCR-related challenges encountered in the
           dataset.}
  \label{fig:ocr_challenges}
\end{figure*}

\begin{comment}
    
\begin{figure}[ht]
\centering
\includegraphics[width=0.95\linewidth]{language_samples.png}
\caption{Example questions from the languages Bulgarian, English, German, Chinese, Croatian, and Hungarian.}
\label{fig:lang_examples}
\end{figure}
\end{comment}

The dataset includes both multiple-choice questions and visual reasoning problems. However, several challenges were observed:

\textbf{OCR-specific Challenges:}
Some items were printed in a language different from their metadata tag, while others lacked standard option labels (\texttt{A–E}) or used a different script problems that confused OCR and downstream prompts (see Figure.~\ref{fig:ocr_challenges}).

\textbf{VLM-specific Challenges:}
Visual-language models often missed important details or made severe misinterpretations. Some image-based questions referenced diagrams that were missing entirely, leading to hallucinated or irrelevant answers. Table \ref{tab:vlm_challenge} shows that the gemini-2.5-flash model has misinterpreted the image saying that the vessel X is from the right ventricle.

\newcolumntype{L}{>{\raggedright\arraybackslash}p{0.16\textwidth}}

\begin{table}[ht]
  \centering
  \caption{VLM mis-interpretation vs.\ our ensemble fix.}
  \label{tab:vlm_challenge}
  \renewcommand{\arraystretch}{1.05}
  \setlength{\tabcolsep}{6pt}

  \begin{tabularx}{\linewidth}{@{} L X @{}}
    \toprule
    \multicolumn{2}{c}{%
      \includegraphics[width=0.60\linewidth]{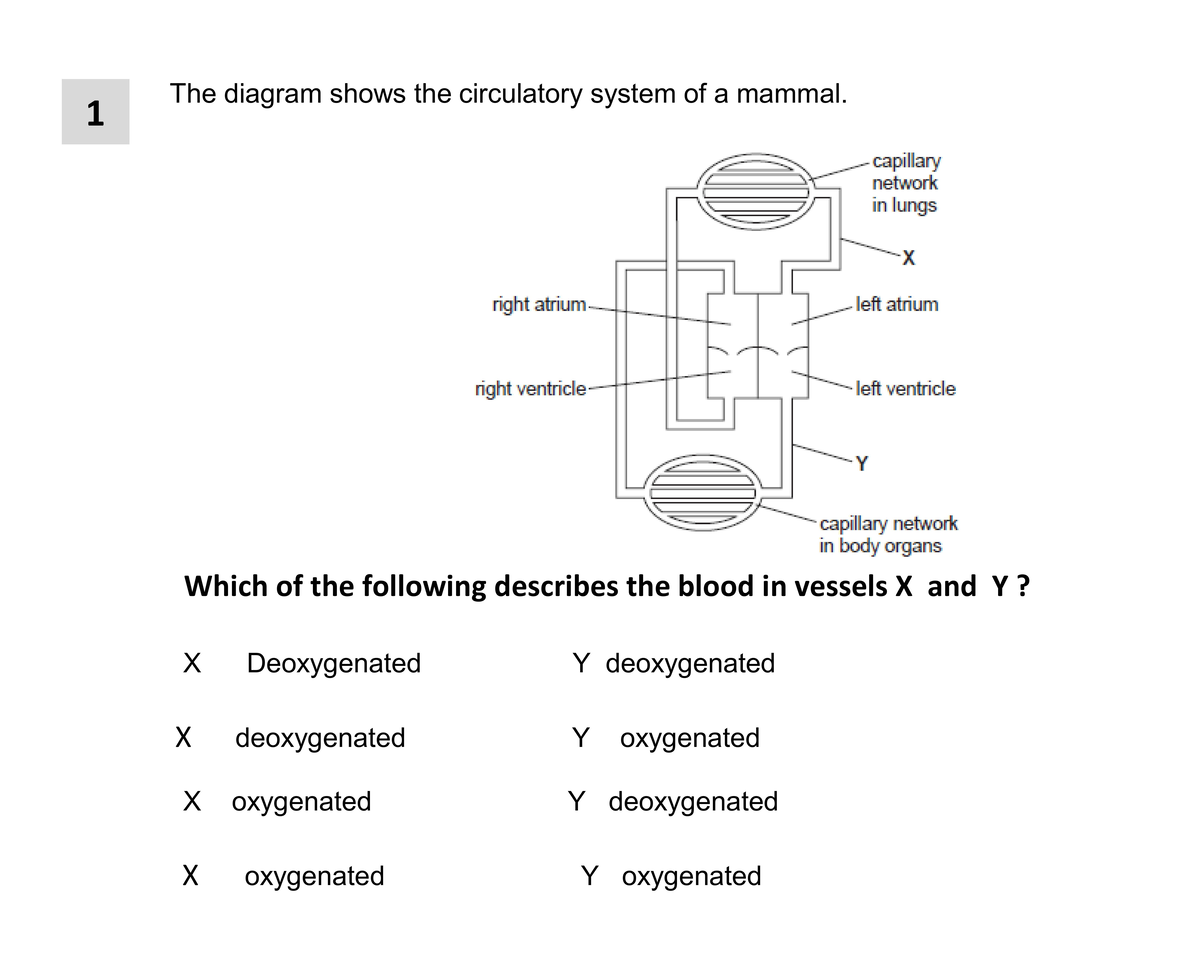}%
    }\\
    \midrule

    % ---------- prompt ----------
    \small\textbf{Prompt} &
    \small Extract the \emph{Question} and \emph{all answer options},
    then provide a detailed, step-by-step description of every key
    visual element. Do not answer the question. \\

    \addlinespace[6pt]   % ← vertical gap
\midrule
    % ---------- single-model failure ----------
    \small\textbf{Gemini 2.5 Flash (VLM)} &
    \small\textit{Description (truncated):} Four-chamber heart; vessel X \emph{from right ventricle to lungs}, vessel Y from left ventricle to body organs.\newline
    \textit{Predicted answer:} \textbf{B} (X deoxygenated, Y oxygenated) — \textit{(incorrect)}.\\

    \addlinespace[6pt]   % ← vertical gap

    % ---------- ensemble success ----------
    \small\textbf{Our Ensemble (VLM)} &
    \small\textit{Description (truncated):} Heart with labelled chambers; vessel X returns \emph{oxygenated} blood from the lung capillary network to the left atrium, vessel Y carries \emph{oxygenated} blood from the left ventricle to body organs\newline
    \textit{Predicted answer:} \textbf{D} (X oxygenated, Y oxygenated) — \textit{(correct)}.\\

    \addlinespace[6pt]
    \midrule
    \small\textbf{Ground Truth} & 
    \small\textit{Answer:} \textbf{D}\\
    \bottomrule
  \end{tabularx}
\end{table}

\textbf{Reasoner-specific Challenges:}
Large Language Models sometimes responded with full sentences or explanations instead of returning a concise choice like ``A'' or ``D,'' which was required by the evaluation format.

\begin{comment}
To address these challenges, we adopted several strategies:

\textbf{OCR \& VLM Pipeline Enhancements:}
We used zero-shot and few-shot prompting to refine visual content interpretation. A describer model generated detailed interpretations of images, which were then verified and corrected using an aggregator model.

\textbf{Multilingual Alignment:}
We curated language mappings and consistency checks to align OCR results with question metadata and ensure correct label extraction.

\textbf{Standardized Output Enforcement:}
Post-processing steps ensured that LLM outputs adhered to the required format (e.g., a single letter choice).
\end{comment}

\begin{table}[htpb]
    \caption{Dataset statistics by language, showing number of subjects, questions, and visual/textual distribution. \cite{das-etal-2024-exams}}
    \begin{center}
    \renewcommand{\arraystretch}{1.3}
    \begin{tabular}{ccccccc}
        \hline
        \textbf{Language} & \textbf{ISO} & \textbf{Family} & \textbf{Grade} & \textbf{\# Subjects} & \textbf{\# Questions} & \textbf{\# Visual Q. / Text Q.} \\
        \hline
        English & en & Germanic & 11, 12 & 4 & 724 & 181 / 543 \\
        Chinese & zh & Sino-Tibetan & 8--12 & 6 & 2,635 & 1,991 / 644 \\
        French & fr & Romance & 12 & 3 & 439 & 50 / 389 \\
        German & de & Germanic & 12 & 5 & 819 & 144 / 675 \\
        Italian & it & Romance & 12 & 6 & 1,645 & 292 / 1,353 \\
        Arabic & ar & Semitic & 4--12 & 6 & 823 & 117 / 706 \\
        Polish & pl & Slavic & 12 & 6 & 2,158 & 72 / 2,086 \\
        Hungarian & hu & Finno-Ugric & 12 & 6 & 3,801 & 495 / 3,306 \\
        Bulgarian & bg & Slavic & 4, 12 & 4 & 2,132 & 435 / 1,697 \\
        Croatian & hr & Slavic & 12 & 6 & 3,969 & 700 / 3,269 \\
        Serbian & sr & Slavic & 12 & 11 & 1,434 & 259 / 1,175 \\
        \hline
    \end{tabular}
    \label{tab:dataset_stats}
    \end{center}
\end{table}

The dataset statistics highlight the diversity of the challenge as shown in Table~\ref{tab:dataset_stats}. For instance, Hungarian and Croatian had over 3{,}800 and 3{,}900 questions respectively, with a high proportion of visual questions. In contrast, English had fewer overall questions but maintained a balance between visual and textual modalities. This linguistic and subject-area diversity posed unique challenges for cross-lingual and multimodal generalization.

The task evaluated over this dataset is:
\begin{itemize}
    \item \textbf{Task 1 – Visual Question Answering (VQA):} Assessing the ability to answer questions based on both images and accompanying text.
\end{itemize}

\begin{figure}[h]
    \centering
    \includegraphics[width=1\linewidth]{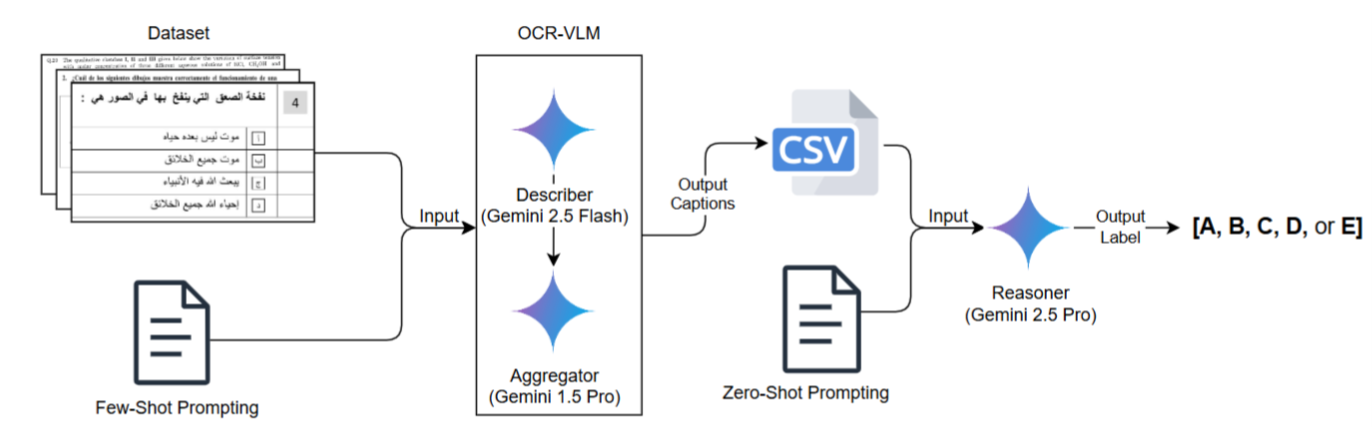}
    \caption{System pipeline: OCR--VLM ensemble (Gemini 2.5 Flash + Gemini 1.5 Pro) produces text for LLM answer selection (Gemini 2.5 Pro).}
    \label{figure:pipeline}
\end{figure}

This task investigates different aspects of multimodal and multilingual reasoning and exposes the weaknesses and strengths of current VLM and LLM systems in handling such richly varied content.

\section{Methodology}
\label{method}

\subsection{Overall Workflow}

As shown in Figure~\ref{figure:pipeline}, our system is a two–stage ensemble pipeline, inspired by recent advances in vision-language and large language models~\cite{zhang2024vision,li2024m4u,gao2025pm4benchparallelmultilingualmultimodal}.  
First, an \textbf{OCR–VLM stage} extracts rich textual descriptions from each
question image; second, a \textbf{Reasoner stage} maps the cleaned text to a final
multiple–choice answer.

\subsection{Stage 1: OCR–VLM Ensemble}

\textit{Gemini 2.5 Flash (describer).}
We employ Gemini 2.5 Flash to generate a detailed natural-language caption of the input image.  A \emph{few–shot} prompt (1~example) is prepended to encourage the model to:
\begin{itemize}
    \item Preserve mathematical symbols and subscripts,
    \item Normalise answer-option markers (``(A)'', ``A.\,'', ``\ding{172}'', \textit{etc.}),
    \item Output in the language inferred from document metadata.
\end{itemize}
Few-shot prompting and multilingual captioning have proven effective in recent VLM research~\cite{zhou2022learning,shi2022languagemodelsmultilingualchainofthought}.

\textit{Gemini 1.5 Pro (aggregator).}
The caption is passed together with the original image to Gemini 1.5 Pro, which acts as a verifier. It is prompted to correct label mismatches, flag missing diagrams (``diagram above'' errors), and translate stray text into the declared language.

\subsection{Stage 2: Reasoner}

Gemini 2.5 Pro receives the caption from each row in the CSV plus a \emph{zero-shot} prompt, following best practices in multilingual reasoning evaluation~\cite{li2024m4u,huang2023m3exam,bi2025reasoning}. We chose Gemini 2.5 Pro over Gemini 2.5 Flash for the final reasoning stage due to its superior performance in complex reasoning tasks and better adherence to strict output formatting requirements~\cite{gemini25comparison,gemini25benchmarks}. While Flash excels in vision-language understanding, Pro demonstrates enhanced logical reasoning capabilities and more reliable response formatting, which are critical for the multiple-choice answer selection task. Gemini 2.5 Pro was selected due to its state-of-the-art performance in Global MMLU (Massive Multitask Language Understanding) with a score of 89.8\%, making it a very reliable choice for this task~\cite{gemini25pro2025}.

\begin{tcolorbox}[colback=gray!5, colframe=black!50, title=Zero-Shot Reasoner Prompt, label=box:llm_prompt,]
You are given a multiple-choice question extracted from an exam.\\
The question description is: \{caption\}\\

Perform the following analysis:\\
1. Carefully read and interpret the full question description provided in the caption.\\
2. Identify the main question being asked.\\
3. Extract all available answer options presented in the description.\\
4. Pay close attention to any data mentioned (tables, diagrams, charts, graphs, \\
\phantom{4. }chemical structures, etc.).\\
5. Analyze all information in context.\\
6. Select the correct answer based solely on your analysis of the provided description.\\

Your final response MUST be ONLY the single letter of the correct answer option ["A", "B", "C", "D", or "E"] in English.\\
Absolutely NO other text, explanation, reasoning, or formatting is allowed in your response. Just the letter.
\end{tcolorbox}

% --------------------------------------------------------------------
\section{Experiments and Results}\label{sec:experiments}
% --------------------------------------------------------------------

\subsection{Experimental Setup}

All submissions were evaluated on the public leaderboard for
\textit{MultimodalReasoning}~\cite{ImageCLEFmultimodalReasoningOverview2025}. 
Accuracy is computed as the fraction of questions for which the system
returned the correct letter (A–E), following the competition’s official evaluation protocol~\cite{OverviewImageCLEF2025}.  
Our system runs the two–stage pipeline described in
Section~\ref{method}:  
Gemini 2.5 Flash\,$\rightarrow$\,Gemini 1.5 Pro for
OCR\,+\,VLM, followed by Gemini 2.5 Pro for reasoning and answer selection.
Unless otherwise stated, ensemble inference uses
{\small\texttt{temperature=1.5 (2.5 Flash)}}, {\small\texttt{1.5 (1.5 Pro)}},
and {\small\texttt{0.2 (2.5 Pro)}}.

\subsection{Performance}

To assess the effectiveness of our approach, we compared our system's accuracy against the organiser-supplied baseline across all supported languages. Table~\ref{tab:leaderboard} summarises the official results, showing the substantial performance gains achieved by our ensemble pipeline. Notably, our system ranked first on the multilingual leaderboard and achieved top ranks in nearly all individual language tracks.

% Table~\ref{tab:leaderboard} contrasts our accuracy with the
% organisers’ baseline for every Language.  
% MSA ranks \textbf{1st} in the multilingual leaderboard and
% in 11 out of 13 languages (Hungarian being the only track we did not
% submit to).

\begin{table}[ht]
\centering
\caption{``Baseline'' is the organizer-supplied reference system. $\Delta$ denotes the absolute accuracy gain.}
\label{tab:leaderboard}
\small
\begin{tabular}{l@{\hspace{6mm}}|@{\hspace{6mm}}c@{\hspace{6mm}}c@{\hspace{6mm}}c@{\hspace{6mm}}|@{\hspace{6mm}}c}
\toprule
\textbf{Language} & \textbf{Baseline} & \textbf{MSA} & \textbf{$\Delta$} & \textbf{Rank} \\
\midrule
Multilingual & 27.01\% & 81.40\% & +54.39\% & 1st \\
Arabic       & 27.03\% & 67.57\% & +40.54\% & 1st \\
Chinese      & 26.78\% & 83.05\% & +56.27\% & 1st \\
German       & 31.01\% & 89.15\% & +58.14\% & 1st \\
Italian      & 24.14\% & 92.12\% & +67.98\% & 1st \\
Spanish      & 31.56\% & 71.98\% & +40.42\% & 1st \\
Urdu         & 30.11\% & 80.67\% & +50.56\% & 1st \\
Serbian      & 23.65\% & 71.43\% & +47.78\% & 1st \\
Croatian     & 27.09\% & 95.07\% & +67.98\% & 1st \\
Polish       & 29.34\% & 82.24\% & +52.90\% & 1st \\
Kazakh       & 27.38\% & 81.48\% & +54.10\% & 1st \\
English      & 24.80\% & 86.52\% & +61.72\% & 2nd \\
Bulgarian    & 24.50\% & 75.00\% & +50.50\% & 3rd \\
\bottomrule
\end{tabular}
\end{table}

\subsection{Ablation Study: Model Architecture and Prompt Engineering}

We conducted a comprehensive ablation study to evaluate the impact of (1) model architecture and scale, (2) multilingual data augmentation, and (3) prompt engineering on multilingual multimodal reasoning performance.

\textbf{Model architecture and multilingual data augmentation.}  
The original English dataset consisted of 377 training and 347 validation questions. To enrich training data with cross-lingual reasoning patterns, we expanded this dataset to 6,841 training and 2,990 validation items by translating questions from 12 other languages into English using Gemini 1.5 Pro. We then fine-tuned three large language models—\textbf{Phi-4} (14B parameters), \textbf{Gemma-3} (12B parameters), and \textbf{Mistral} (7B parameters)—on both the original and expanded datasets. Additionally, Gemini 2.5 Flash was evaluated in a zero-shot setting via API to justify its selection as the vision-language component in our system.  

Table~\ref{tab:ablation_models} summarizes the results. The findings reveal that multilingual augmentation significantly improves performance for larger models: Phi-4 and Gemma-3 gained +19.63 and +19.96 percentage points, respectively. However, Mistral (7B) showed only minimal benefit (+0.74 pp), suggesting insufficient capacity for complex cross-lingual reasoning. Gemini 2.5 Flash achieved a substantial gain of +12.79 pp, from 66.86\% on the unexpanded dataset to 79.65\% on the expanded dataset, outperforming all other models and validating its role in our system.  

\begin{table}[htbp]
    \centering
    \caption{Model ablation results on unexpanded and expanded datasets. Gemini 2.5 Flash was evaluated zero-shot via API (not fine-tuned).}
    \label{tab:ablation_models}
    \renewcommand{\arraystretch}{1.3}
    \begin{tabular}{lccc}
        \hline
        \multirow{2}{*}{\textbf{Model}} & \textbf{Parameters} & \multicolumn{2}{c}{\textbf{Accuracy (\%)}} \\
        \cline{3-4}
        & \textbf{(B)} & \textbf{Unexpanded Dataset} & \textbf{Expanded Dataset} \\
        \hline
        Gemini 2.5 Flash* & --  & 66.86       & \textbf{79.65} \\
        Phi-4             & 14  & 36.02       & 55.65 \\
        Gemma-3           & 12  & 23.92       & 43.88 \\
        Mistral           & 7   & 27.09       & 27.83 \\
        \hline
    \end{tabular}
\end{table}

\textbf{Prompt engineering.}  
We further analyzed the role of prompt design by testing different prompting strategies on the English validation set. Switching from a verbose descriptive prompt to a strict “answer-letter-only” instruction boosted Gemini Flash accuracy from 55.9\% to 57.1\%. Replacing Flash with Gemini 1.5 Pro under the same prompt further increased accuracy to 61.7\%, suggesting that larger models can exploit strict prompts more effectively (Table~\ref{tab:prompt_results}).  

\begin{table}[htbp]
    \centering
    \caption{Prompt-ablation results on the English validation split for the Reasoner stage.}
    \label{tab:prompt_results}
    \renewcommand{\arraystretch}{1.2}
    \begin{tabular}{lccc}
        \hline
        \textbf{Model}  & \textbf{Prompt Style}       & \textbf{Shots} & \textbf{Accuracy (\%)} \\
        \hline
        2.5 Flash       & long descriptive           & few            & 55.91 \\
        2.5 Flash       & strict letter-only         & few            & 57.06 \\
        1.5 Pro         & strict letter-only         & few            & \textbf{61.67} \\
        \hline
    \end{tabular}
\end{table}

These results emphasize the importance of both architectural choices and precise prompt design in building effective multilingual multimodal reasoning systems.

\subsection{Discussion}

Our experiments highlight several key insights:  

First, the ablation study demonstrates that both model scale and multilingual data augmentation are critical for achieving high reasoning accuracy. Larger models such as Phi-4 and Gemma-3 benefited substantially from training on the expanded dataset, whereas Mistral (7B) showed minimal improvement, indicating limited capacity for complex cross-lingual reasoning. Gemini 2.5 Flash, even without fine-tuning, consistently outperformed these models, underscoring the value of large-scale pretraining and advanced multimodal capabilities.

Second, prompt engineering played a pivotal role in optimizing performance. Strict output constraints, which prohibited explanatory text and enforced concise letter-only answers, reduced failure cases caused by “overflow” responses. Gemini 1.5 Pro exploited this prompt design more effectively than Gemini 2.5 Flash, suggesting a synergy between prompt quality and model capacity.

Finally, our findings reinforce the design choices of our ensemble system. By combining lightweight OCR–VLM components for vision-language understanding with a reasoning-optimized LLM, we achieved state-of-the-art performance in multilingual educational QA tasks. 

\section{Conclusion}

In this paper, we presented a robust ensemble-based approach for multilingual multimodal reasoning, integrating Gemini 2.5 Flash and Gemini 1.5 Pro for vision-language tasks with Gemini 2.5 Pro as the final reasoner. Through careful prompt engineering and strict output normalization, our system achieved state-of-the-art performance on the ImageCLEF 2025 Multimodal Reasoning leaderboard, ranking first overall and securing the top position in 11 out of 13 language-specific tracks. The ablation study highlighted the importance of model architecture, multilingual data augmentation, and precise prompt design, demonstrating significant accuracy gains and validating the choice of Gemini 2.5 Flash as the backbone for our system, especially in handling languages with complex scripts. 

Our findings underscore that combining lightweight, well-calibrated OCR–VLM pipelines with targeted prompt strategies can outperform heavier end-to-end models, particularly in high-stakes educational scenarios requiring reliable automatic grading. Nonetheless, challenges remain, especially regarding the handling of ambiguous diagrams and enforcing strict output formats in low-resource languages. Future work will explore reinforcement learning for format adherence, enhanced diagram processing, and further augmentation for underrepresented languages. 

Overall, our results confirm that prompt-centric system design and ensemble modeling represent a powerful paradigm for advancing multilingual and multimodal question answering~\cite{ImageCLEFmultimodalReasoningOverview2025,OverviewImageCLEF2025,li2024m4u,huang2023m3exam,bi2025reasoning}.
%%
%% Define the bibliography file to be used
\bibliography{sample-ceur}


\end{document}